# Convolutional Neural Network for emotion recognition to assist psychiatrists and psychologists during the COVID-19 pandemic: experts' opinion

Hugo Mitre-Hernández[a*], Rodolfo Ferro-Perez[a], Francisco Gonzalez-Hernandez[b]

[a]Department of computer science, Mathematics Research Center (CIMAT), Zacatecas, México;
[b]Tecnologico Nacional de México, Campus Culiacán, México.

*Hugo Mitre-Hernandez, hmitre@cimat.mx , ORCID: 0000-0003-2840-3998. CIMAT Building, Quantum: ciudad del conocimiento, Zacatecas, Zacatecas, Mexico. C.P. 98160.

**Abstract**: A web application with real-time emotion recognition for psychologists and psychiatrists is presented. Mental health effects during COVID-19 quarantine need to be handled because society is being emotionally impacted. The human micro-expressions can describe genuine emotions that can be captured by Convolutional Neural Networks (CNN) models. But the challenge is to implement it under the poor performance of a part of society's computers and the low speed of internet connection, i.e., improve the computational efficiency and reduce the data transfer. To validate the computational efficiency premise, we compare CNN architectures results, collecting the floating-point operations per second (FLOPS), the Number of Parameters (NP) and accuracy from the MobileNet, PeleeNet, Extended Deep Neural Network (EDNN), Inception- Based Deep Neural Network (IDNN) and our proposed Residual mobile-based Network model (ResmoNet). Also, we compare the trained models' results in terms of Main Memory Utilization (MMU) and Response Time to complete the Emotion (RTE) recognition. Besides, we design a data transfer that includes the raw data of emotions and the basic patient' information. The web application was evaluated with the System Usability Scale (SUS) and a utility questionnaire by psychologists and psychiatrists. ResmoNet model generated the most reduced NP, FLOPS, and MMU results, only EDNN overcomes ResmoNet in 0.01sec in RTE. The optimizations to our model impacted the accuracy, therefore IDNN and EDNN are 0.02 and 0.05 more accurate than our model respectively. Finally, according to the psychologists and psychiatrists, the web application has good usability (73.8 of 100) and utility (3.94 of 5).

**1. Introduction**

Emotions "*are a process, a particular kind of automatic appraisal influenced by our evolutionary and personal past, in which we sense that something important to our welfare is occurring, and a set of psychological changes and emotional behaviors begins to deal with the situation*" [1]. There are several negative emotional behaviors in the society evoked by certain situations during the COVID-19 pandemic.

Chinese children and adolescents are experiencing fear due to interrogatories about the COVID-19 pandemic [2]. Also, the unpleasant experience of quarantine is generating anger in society [3], sadness, and disgust [4]. Emotions are mainly influenced by depression, anxiety, and stress. *Anger*, *fear*, *sadness,* and *happiness* are evoked emotions during the clinical disorders of depression, anxiety, and mixed anxiety-depression. Also, *sadness* and *disgust* can be elevated in the depressed and mixed disorders, whereas increased levels of anger and fear, and decreased levels of *happiness* [4], indicating a stress situation, in such a way, post-traumatic stress disorder was reported in China, Spain, Italy, Iran, the US, Turkey, Nepal, and Denmark [5]. Also, sadness and fear are evoked in social media, and their intensity fluctuates over time [6]. The clinical disorders and its origins need to be explored.

Much of society's emotions are being affected by the COVID-19 pandemic. Not only infected people are affected, but also children, parents, healthcare workers, and more. Children's mental disturbance is tied to parent's stress [7], and health care workers have higher levels of psychiatric symptoms [8]. We consider much of society is emotionally affected, for this reason, psychological and psychiatric intervention is urgently needed.

The evidence-based treatments for an individual with emotional disorders are basic for its empirical validation [9]. That is why emotional records are indispensable as part of such evidence. Is well known that movements of facial muscles are associated with emotions [10]. Spontaneous emotions are typically expressed through facial micro-expressions. Micro-expressions have been a growing interest in computer vision and artificial intelligence studies to automate emotion recognition; the researchers name it Facial Emotion Recognition (FER) and its data can be used as emotional evidence for the treatments of psychologists and psychiatrists.

There is an imminent risk of contagion between patient-psychologist or patient- psychiatrist in consulting rooms. Taking advantage of the fact that 70.1% of the population aged 6 and over in Mexico has access to the internet [11], web applications can mitigate this risk. But the major problem in México and other countries is the poor performance of computers in a major part of society and the low speed of internet connection in several regions. For these reasons, we propose a web application to record real-time emotions in a patient's card (personal and treatment information) developed with the following considerations: the CNN model must generate fewer processor' operations using less primary memory and reducing time to emotions recognition than other models to achieve the performance needed, data transfer size must be very small to works in low bandwidth internet connections. Also, our proposal needs to be evaluated by psychologists and psychiatrists with many years of experience and patients treated due to the COVID-19 circumstances.

Comparing the results of our Residual mobile-based Network (ResmoNet) model with the best model. It was obtained that ResmoNet generated 115,976 number of parameters less than MobileNet, 243,901 floating-point operations per second (FLOPS) less than MobileNet, and 5% less accuracy than EDNN (95%). Moreover, ResmoNet used less main memory utilization than any model, only EDNN overcomes ResmoNet in 0.01 seconds for the time to complete the emotion recognition. For data transfer, the patient's card and raw emotional data have 2 kb with a UTF-8 encoding approximately. Finally, according to the psychologists and psychiatrists, the web application has good usability (73.8 of 100) and utility (3.94 of 5).

## 2. Method

We present a background of CNN optimized models, the proposed CNN model with its the preprocessing and training stages and the web application for psychiatrists and psychologists. Also, the methods to compare the IDNN, EDNN, MobileNet and to evaluate the usability and utility the Web application.

*2.1 Optimized CNN Models*

In deep-learning-based FER approaches, the Deep Convolutional Neural Networks (Deep CNNs) have a large performance improvement in comparison to conventional unsupervised approaches of machine learning [12–14]. In general, the function of CNNs models receive the face image, it is convolved through a filter collection in the convolution layers to create the feature map, this is combined to fully connected networks, and then face expression is recognized into a particular class as the output of the softmax algorithm [15]. Studies of CNN models need to be analyzed to design an optimized model.

For CNN models effectiveness, the Residual Networks (ResNet) are easier to optimize and can improve the accuracy from considerably increased depth [16]. ResNet models are often implemented with two or three layers skips that contain non-linearities (ReLU, Refined Linear Units) and batch normalization [17] or concatenate --e.g, the EDNN model [18] in between. Another strategy to reduce the floating-point operations is the depthwise separable convolution defined in MobileNet [19], which deals with the depth dimension --the number of channels. A depthwise separable convolution works with kernels that cannot be "factored" into two smaller kernels, therefore, it splits the kernel into 2 separate kernels that do two convolutions: the depthwise convolution and the pointwise convolution. Inception module architecture [20] deploys multiple convolutions with multiple filters and pooling layers simultaneously in parallel within the same layer --the inception layer. Mollahosseini et al. [21] confirmed in its proposal of Inception-Based Deep Neural Network (IDNN), the use of this module can increase the depth and width of the architecture while preserves the computational cost constant.

MobileNet model was built primarily from depthwise separable convolutions used in inception models [22] to reduce the computational cost in the first few layers. Depthwise separable convolution deals with the depth dimension, i.e. the number of channels. A depthwise separable convolution works with kernels that cannot be "factored" into two smaller kernels. Therefore, it splits the kernel into two separate kernels that do two convolutions: the depthwise convolution and the pointwise convolution (a simple 1x1 convolution). In the comparative study of [19], MobileNet achieved a 79.4% of accuracy with 5.60 million less Mult-Adds (computation) compared to FaceNet [23] who reached 83% accuracy. The MobileNet's architecture generates fewer Mult-Adds, sacrificing accuracy for resource-limited devices.

PeleeNet [17] model is a variant of the DenseNet architecture that follows its connectivity pattern [24]. Looking to optimize memory and computational cost, its architecture was based on (i) the two-way dense layer of GoogLeNet [25]; (ii) the inception-v4 [26] and the deeply supervised object detectors [27] to design a cost-efficient stem block before the first dense layer; and (iii) bath normalization [22] for post-activation in the composite function. The major improvements of speed and accuracy in PeleeNet were the adjustments to (i) the feature map selection based on Single Shot MultiBox Detector [28] in 5 scale features: 19x19, 10x10, 5x5, 3x3, and 1x1 (excluding 38 x 38 feature map); (ii) the residual prediction block [16] for each feature map; and (iii) a 1x1 convolutional kernels for prediction, this reduced 21.5 % of computational cost and the model size (number of parameters). PeleeNet and MobileNetV2 were evaluated on NVIDIA TX2 embedded platform processing 100 pictures with 1 batch size, even though MobileNetV2 achieves high accuracy with 300 FLOPS, the speed of PeleeNet was better than MobileNet with 569 FLOPS.

The EDNN model was designed for facial emotion recognition [18]. For face detection authors proposed a deep convolution neural network composed of six convolution layers, two blocks of deep residual learning, after each convolution layer. Also, deep residual blocks are implemented after the second and fourth convolution, two fully connected layers, each with a ReLU activation function, and dropout for the training layer. The combination of fully connected layers and residual blocks could improve the overall result. EDNN obtained the accuracy of 93.24 % trained with the Extended Cohn–Kanade dataset (CK+) [29] for emotion recognition of sad, happy, surprise angry, disgust, and fear, surpassing other models of various authors that presented a lower accuracy at 92.71 % [30], 92.35 % [31], 92.74 % [32], and 92.73 % [33].

IDNN [21] model was inspired by GoogLeNet [20] and AlexNet [34], it consists of two CNN modules –a convolution layer by a max-pooling layer– using the ReLU activation function Refined Linear Units to avoid the vanishing gradient problem. After these two modules, the network-in-network architecture and two inception modules were applied –made up of 1x1, 3x3, and 5x5 convolution layers using ReLU in parallel. The inception layers are concatenated as output and, finally, two fully connected layers as the classifying layers. The trained model of IDNN with the CK+ dataset obtained the average accuracy of 97.8 % in comparison with other works at 88.5 % [25], 92 % [35], 92.4 % [36] and 93.6 % [37]. The authors create their proposal based on the inception layer to improve the recognition of local features such as the eye and mouth, expressions that can describe an emotion. Also, they applied the network-in-network theory [38] to enhance the local feature performance, increase the global pooling performance and reduce prone to overfitting.

Our CNN model was based on several optimizations mentioned before, mainly on stem, mobile and residual block structures, small kernel size, among other optimization practices.

*2.2 Residual mobile-based Network*

Our proposal ResmoNet contains a stem block, which is fed with our input images, followed by the composition of two networks: a mobile network containing one mobile block, and a residual network containing a residual block and a transition block. After these blocks, a dense block of neurons is added before the output layer. Motivated by [17], we implemented a Stem Block before the mobile section of our proposal. The structure of the Stem Block can be seen in Figure 1.

**Fig. 1**

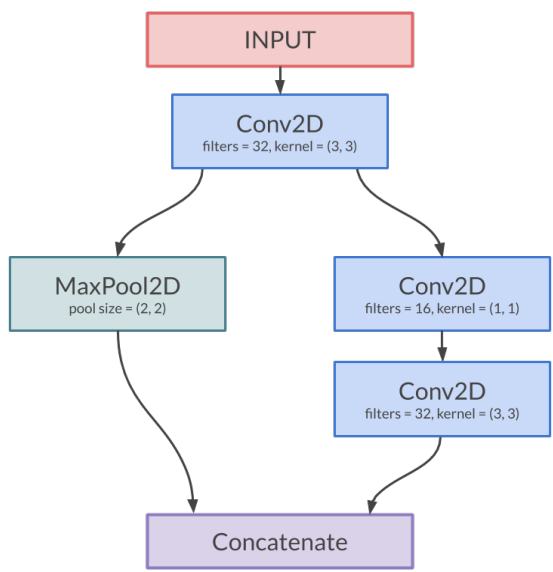

Stem Block structure

The Mobile Network that follows the Stem Block can be composed of a sequence of *m* Mobile Blocks by defining the parameter *m*, which is a mobile-depth parameter (it specifies the number of Mobile Blocks). Our proposal contains a single Mobile Block ($m=1$). The Mobile Block is implemented as defined in [19], which is structured by the following sequence: a depthwise 2D convolution with batch normalization and ReLU as the activation function, a 2D pointwise convolution with batch normalization and ReLU as the activation function, and finally an average pooling.

The Residual Block that follows the Mobile Block can be composed of a sequence of *r* Residual Blocks by defining the parameter *r*, which is a residual-depth parameter (it specifies the number of Residual Blocks) and a Transition Block. It contains a single Residual Block ($r=1$). The structure of the Residual Block can be seen in Figure 2. The Transition Block is structured as follows: it contains a 2D convolution with batch normalization and ReLU as the activation function and finally an average pooling.

**Fig. 2**

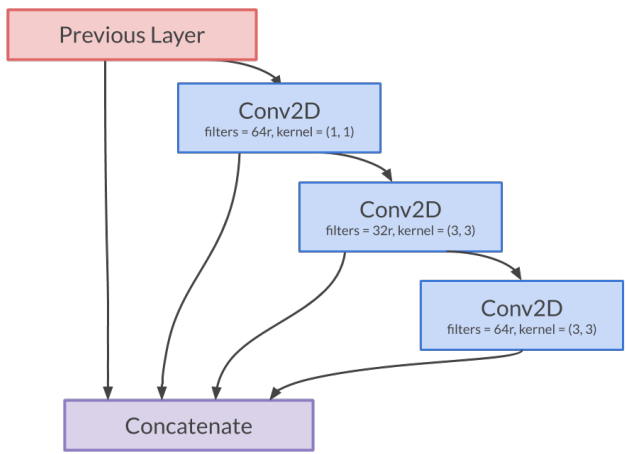

Residual Block structure.

Finally, a Dense Block of two layers with dropout is added. A complete overview of the model architecture is shown in Figure 3.

**Fig. 3**

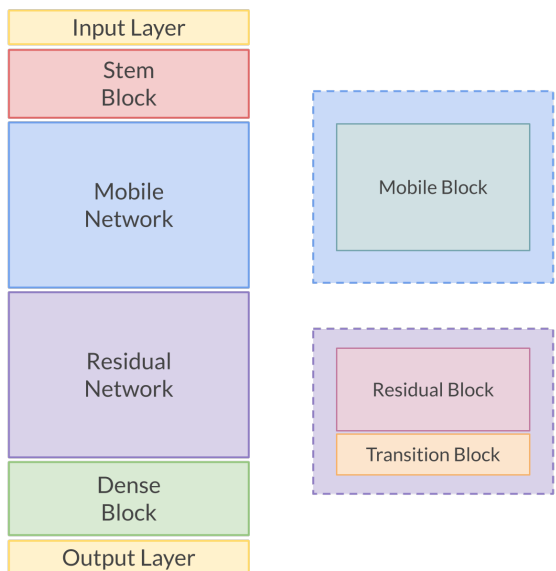

Overview of the ResMoNet model architecture.

ResmoNet was built using the Keras library [39] on top of Google's TensorFlow [40] framework. The versions used were 2.2.4 on Keras and 2.0.0 on TensorFlow. For image preprocessing, we used OpenCV 3.4.3 with Python.

The training of our proposed neural network was done on a machine with the following hardware specifications: GPU: 1×Tesla K80 with CUDA 10.1; CPU: 1×Single core hyperthreaded (i.e. 1 core, 2 threads) Intel(R)Xeon(R) CPU @ 2.30GHz, 45 MB Cache; RAM: ~12 GB available; Disk: ~310 GB available.

Once that our proposed model was trained, it was deployed, tested, and measured on a Raspberry Pi 3 Model B with the same software specifications as the previously mentioned for training, and with the following hardware specifications: CPU: 4×ARM Cortex-A53, 1.2GHz; GPU: Broadcom VideoCore IV; RAM: 1GB LPDDR2 (900 MHz); Disk: ~32 GB available.

On this deployment, the execution time response (*ETR*) and the main memory utilization (*MMU*) were measured as computational resources for the evaluation of the modeling production.

*2.3 Dataset, Preprocessing, and Training*

The Radboud Faces Database (RaFD) is a high-quality faces database containing pictures of 67 models (including Caucasian males and females, Caucasian children, both boys and girls, and Moroccan Dutch males) displaying 8 emotional expressions (accordingly to the Facial Action Coding System): Anger, disgust, fear, happiness, sadness, surprise, contempt, and neutral. Each emotion was shown with three different gaze directions [41]. To train ResmoNet, the images from the RaFD with front faces were used, including the images with the view to the front and the sides.

To generate the set of ROIs that will be split into the input images to train and test our model, we preprocessed raw images from the RaFD dataset following the steps shown in Figure 4. Each step of the whole preprocess consists of the following sub-processes:

- Face detection on images (see Figure 4A). This process consists of the detection of ROIs inside the images that contain the face. For this, the default frontal face Haar cascades from OpenCV were used and the (*x*, *y*) coordinates of the bounding box around the face were extracted.
- Face cropping from images (see Figure 4B). Once that the face is detected on the image, we use the extracted (*x*, *y*) coordinates of the bounding box to crop the detected face and generate a squared ROI.
- Cropped face resizing (see Figure 4C). This process consists of resizing the ROI (cropped image) into the defined input size for the model (224x224x3), so regarding the sizes for different faces detected, we standardize the input size.

**Fig. 4**

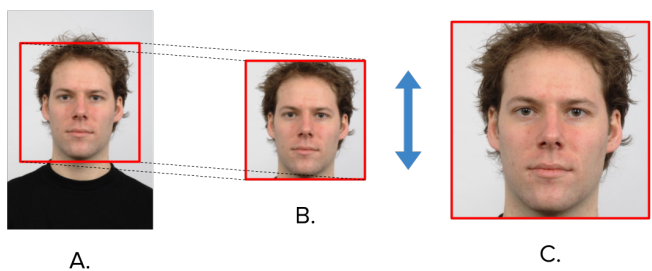

(a) The face is detected from the original image. (b) The detected face is cropped into a new image. (c) The cropped face is resized to the input size used to feed the neural network.

Once that all the images were cropped and resized to have only 224x224x3 sized images of faces, we proceeded to augment data. The augmented data consists of the following new images:

- The sub-section of the top-left corner, removing the remaining border of the bottom-right section to create a sub-image with size 186x186x3 which was also resized to 224x224x3. This process was repeated for the top-right corner, the bottom-left corner, and the bottom-right corner, removing the remaining borders to create a sub-image with size 186x186x3 which were also resized to 224x224x3.

- The sub-section of the center of the image, removing the remaining border around the original ROI to create a sub-image with size 148x148x3 which was also resized to 224x224x3.
- The flipped image around the *Y*-axis of the original image.
- Each of the previous processes for data augmentation for the flipped image.

An example of images generated in the data augmentation process from a single image from the Radboud Faces Database can be seen in Figure 5.

**Fig. 5**

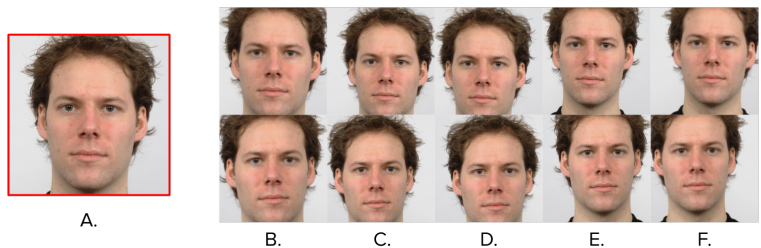

A.   B.   C.   D.   E.   F.

(a) The cropped face from the original image. (b) The sub-section of the center of the cropped image (up) and its horizontally flipped image (down). (c) The sub-section of the top-right corner (up) and its horizontally flipped image (down). (d) The sub-section of the top-left corner (up) and its horizontally flipped image (down). (e) The sub-section of the bottom-right corner (up) and its horizontally flipped image (down). (f) The sub-section of the bottom-left corner (up) and its horizontally flipped image (down).

The preprocessed RafD dataset was split into training (80 %) and testing (20 %) subsets of images. With the training subset, the data augmentation process was applied. Our proposed model, along with the other models for comparison were trained 150 epochs with a batch size of 128 in the previously specified GPU environment. To maintain consistency in the experiment replica for the different network architectures, we set a random seed, so the same images for training and testing were used on each network.

*2.4 Evaluation method of the DNN model*

We performed a comparison of IDNN, EDNN, 1.0 MobileNet, 0.75 MobileNet, and ResMoNet models using computational efficiency measures. Let us define *Computational Efficiency* as the properties of an algorithm/software which relate to the amount of computational resources used by the algorithm/software [42]. The comparison of architectures and trained models in terms of computational properties allow us to know its efficiency for resource-limited devices.

To measure the computational efficiency of the models' architecture, we used the *number of parameters* (*NP*) in a given layer as the count of "learnable" elements in that layer. This means the number of elements to be optimized. Also, the floating-point operations per second (*FLOPS*) is a measure of computer performance, which requires floating-point calculations multiply-accumulate (*Mult-Add*) operations. The *Mult-Add* operation is a common step that computes the product of two numbers and adds that product to an accumulator. Finally, the classification *accuracy* is obtained.

Other measures were collected from the trained model installed in the Raspberry Pi 3 Model B. The *Response Time to complete the Emotion recognition* (*RTE*) [43] is calculated as the average of time records elapsed between the start and completion of a task of emotion recognition; five executions of a minute in each trained model give us the time records. The *Main Memory Utilization* (*MMU*) [44] is measured as the amount of main memory used during trained model execution.

*2.5 Web application*

The web application was built using diverse libraries on the language programming Python, we use Flask, a web framework to process requests from the Internet and Dash with Plotly for generating real-time plots on the web. The web application consists of three main pages for user interaction (see Figure 6).

**Fig. 6**

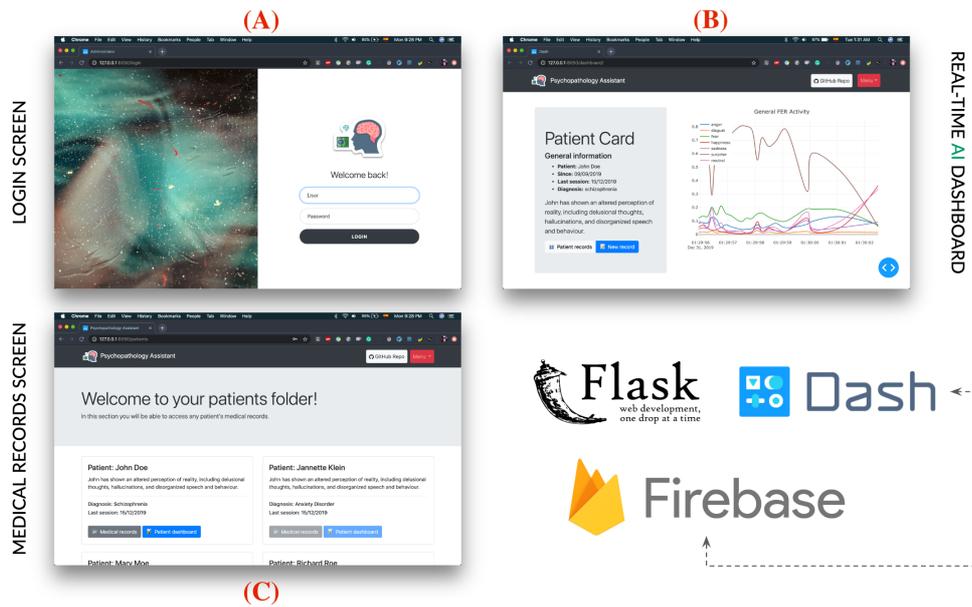

Web application. (a) a login page, which provides access to the psychologist or psychiatrist; (b) A medical records page, which contains medical records for each patient, it displays information on plots of emotions (anger, disgust, fear, happiness, sadness, surprise, and neutral) generated in the web application; (c) A register medical activity page to add new activities in real-time; and the main technologies used.

On the other hand, in the background, the web application continuously consumes a real-time database hosted on Firebase technology, a technology to keep data on the cloud. To saving full histories of interaction from the web application. This database can be downloaded for additional reviews, and it can be filtered in specific periods of any session.

*2.6 Evaluation method of the application web*

*Usability* and *utility* of the web application were evaluated with the experts' support (see the questionnaires in Appendix 1). We consider an expert a psychologist or psychiatrist with more than 5 years of experience, moreover, experts should have cared for patients emotionally affected by the circumstances generated by the COVID-19 pandemic.

We presented a video web application used by the experts. For the usability evaluation, we used the System Usability Scale (SUS) questionnaire [45]. To calculate the SUS score for each expert, we follow the next steps:
1. Convert the scale into numbers for each of the 10 questions
    - Strongly Disagree: 1 point
    - Disagree: 2 points
    - Neither agree nor disagree: 3 points
    - Agree: 4 points
    - Strongly Agree: 5 points
2. Calculate the SUS score:
    - $X$ = Sum of the points for all odd-numbered questions – 5. Odd – questions 1, 3, 5, 7, and 9
    - $Y$ = 25 – Sum of the points for all even-numbered questions. Even – questions 2, 4, 6, 8, and 10
    - *SUS score* = $(X + Y) * 2.5$
3. Finally, *E* is calculated as the average of the SUS scores' experts.

The AVG of SUS scores' experts (*E*) can be evaluated under the following criteria:
- 80.3 is Excellent
- Between 68 – 80.3 is Good
- 68 is Okay
- Between 51 – 68 is Poor
- 51 or less is Awful

Also, inspired by web application utility, we define the utility questionnaire based on this study [46]. For all answers, we used the five-level Likert scale (1 Strongly disagree, 2 Disagree, 3 Neither agree nor disagree, 4 Agree, and 5 Strongly agree). Finally, two opened questions about the patients and web application.

## 3. Results and Discussions

*3.1 CNN Models*

The Table I presents the results of the computational efficiency of the models' architecture (*NP*, *FLOPS*, *Mul-Add* operations, and *accuracy*) and the trained model installed (*RTE* and *MMU*). Resmonet presented the smallest size in architecture, and therefore the minor number of *NP* and *mult-add* operations than IDNN, EDNN, 1.0 MobilNet, 0.75 MobileNet, and PeleeNet models. We consider that concatenation and convolutional blocks (dense blocks) could increase the number of mult-add operations in general. Also, ResmoNet obtained a 90% of *accuracy*, only the EDNN and the IDNN models surpassed our model by 5% and 2% correspondingly. In practice, shallower networks take less time being compiled in our testing portable device. Similarly, the prediction task is more efficient in shallower networks. Deeper networks tend to have a longer *RTE* and more *MMU*. For these reasons, ResmoNet presented the best result in *MMU*, and a difference of 0.01 seconds in *RTE* with the EDNN model.

Table I. Comparison of computational resources and accuracy of different models.

| Model | NP | Mult-Add Ops. | Accuracy | RTE | MMU |
|---|---|---|---|---|---|
| IDNN | 16,158,790 | 32,302,489 | 0.92 | 0.33 sec. | 296.45 MB |
| EDNN | 4,621,638 | 9,235,929 | 0.95 | 0.15 sec. | 245.86 MB |
| 1.0 MobileNet | 3,235,014 | 6,481,263 | 0.55 | 1.17 sec. | 282.32 MB |
| 0.75 MobileNet | 1,837,590 | 3,683,679 | 0.48 | 0.99 sec. | 274.43 MB |
| PeleeNet | 2,123,502 | 4,239,183 | 0.84 | 0.44 sec. | 457.42 MB |
| ResMoNet | 1,721,614 | 3,439,778 | 0.90 | 0.16 sec. | 235.62 MB |

The Figures 7 and 8 show the training history –for train and test sets– of our model in 150 epochs; the resulted accuracy of the training and testing sets show a remarkable closeness as the epochs increase, it can be seen along with the model training that as the accuracy on the training set increases, the accuracy on the testing set behaves similarly. This also happens with the loss function on both sets.

**Fig. 7**

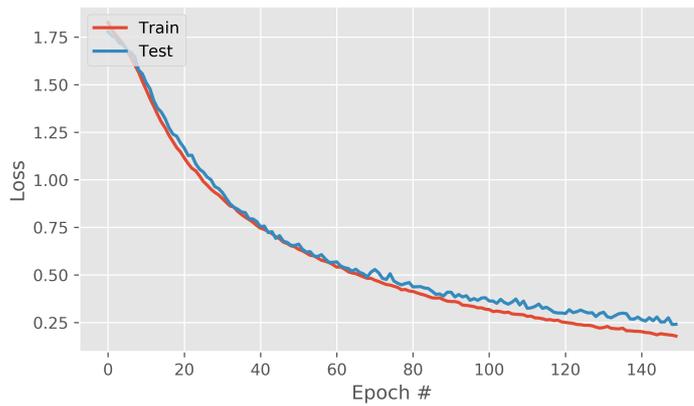

The training graph contains the loss during the model training.

**Fig. 8**

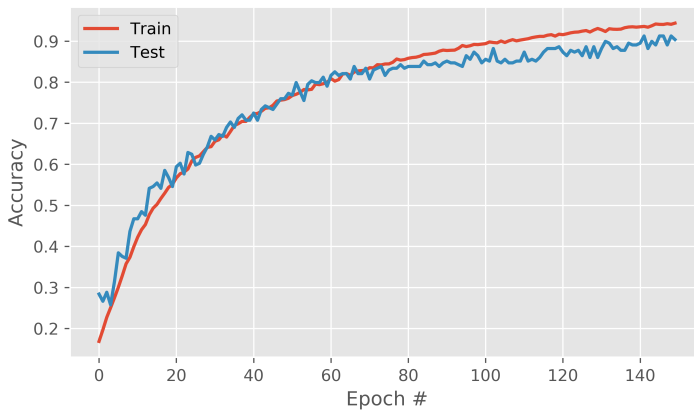

The training graph contains the accuracy during the model training.

The confusion matrix of ResMoNet model on the testing set is shown in Figure 9. It can be seen in the confusion matrix that neutral expression is the most difficult state to recognize correctly, and tends to be confused with a sadness state with 13%; also, surprise with fear can be confused with 11%.

**Fig. 9**

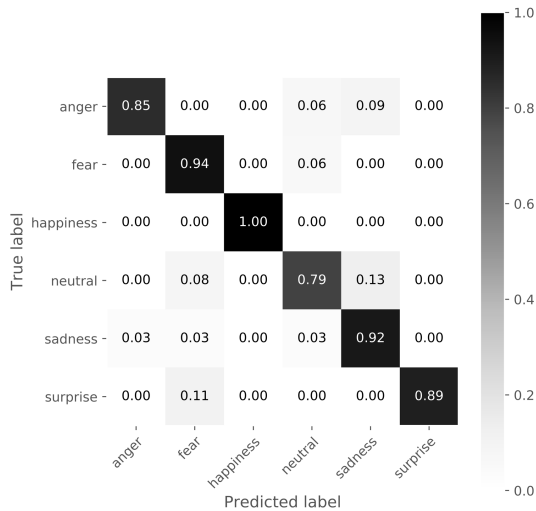

Figure 9. The confusion matrix of the testing data.

3.2 Web application

We obtained valuable information of five adult experts (*M* = 29.8 age, *SD* age =4.7 age), all women with more than 5 years of experience (*M*=11.6 years, *SD*=5.23 years), and all with a postgraduate degree (see Table II).
We consider that Years of Experience (YE) and the number of Patients Treated (PT) of two experts have a different impact on results, e.g., the qualitative data from an expert with 23 *YE* and 260 *PT* has a greater impact on usability and utility results than an expert with 6 *YE* and 21 *PT*. For this reason, as can be seen in Table II, we calculated the *YE* pondered as the years of experience of an expert divided by the experts' total years of experience. Also, the *PT* pondered as the patients treated of an expert divided by the experts' total patients treated. Finally, *YE* and *PT* equally matter, therefore we calculate a weight (*W*) as the average of YE and PT pondered. The real impact on the expert score value is obtained by multiplying the *W* by the final usability or utility score, e.g., if we calculate *W * AVG utility of the expert one, 0.40 * 3.5, the result is 1.39*.

Table II. Experts' information and its qualitative ponderation.

| Expert (*E*) | Years of Experience (*YE*) | Age | Num. of Patients treated (*PT*) | Education degrees | *YE* pondered | *PT* pondered | Weight (*W*) |
|---|---|---|---|---|---|---|---|
| E1 | 23 | 42 | 260 | Bsc. in Psychology, Ms. in family therapy. | 0.34 | 0.46 | 0.40 |
| E2 | 21 | 43 | 86 | Bsc. in Psychology, PhD. in psychology. | 0.31 | 0.15 | 0.23 |
| E3 | 10 | 45 | 3 | Bsc. in Psychology, Ms. in learning difficulties. | 0.15 | 0.01 | 0.08 |
| E4 | 8 | 37 | 200 | Bsc. in Psychiatry, Specialty in Paidopsychiatry. | 0.12 | 0.35 | 0.23 |
| E5 | 6 | 32 | 21 | Bsc. in Psychology, Ms. in Psychology. | 0.09 | 0.04 | 0.06 |

Table III. Usability and utility questionnaires results.

| | # Question | E1 | E2 | E3 | E4 | E5 |
|---|---|---|---|---|---|---|
| Usability (SUS) | 1 | 4 | 4 | 3 | 5 | 4 |
| | 2 | 3 | 1 | 1 | 2 | 3 |
| | 3 | 5 | 2 | 5 | 4 | 3 |
| | 4 | 2 | 5 | 2 | 3 | 2 |
| | 5 | 5 | 5 | 5 | 5 | 4 |
| | 6 | 3 | 2 | 1 | 4 | 3 |
| | 7 | 4 | 5 | 4 | 5 | 5 |
| | 8 | 2 | 1 | 1 | 4 | 1 |
| | 9 | 4 | 4 | 3 | 5 | 2 |
| | 10 | 2 | 3 | 2 | 1 | 3 |
| Utility | 11 | 4 | 5 | 1 | 4 | 4 |
| | 12 | 2 | 4 | 4 | 4 | 4 |
| | 13 | 4 | 5 | 3 | 4 | 4 |
| | 14 | 4 | 5 | 3 | 5 | 4 |
| | Total Odd | 17 | 15 | 15 | 19 | 13 |
| | Total Even | 13 | 13 | 18 | 11 | 13 |
| | SUS Score | 75 | 70 | 82.5 | 75 | 65 |
| | *W usability* | 29.79 | 16.09 | 6.28 | 17.57 | 4.07 |
| | AVG utility | 3.5 | 4.75 | 2.75 | 4.25 | 4 |
| | *W utility* | 1.39 | 1.09 | 0.21 | 1.00 | 0.25 |
| | Total *W* usability | 73.8 | | | | |
| | Total *W* utility | 3.94 | | | | |

*As shown in Table III, the experts considered that the web application has a good utility (3.94 of 5) and usability (73.8 of 100). Also, some experts share its critics:*

- E1. "*It can be a complementary tool for diagnostic processes. Prior to the intervention to solve the problem.*"
    - o Our answer: Exactly, the purpose of the web application, not to replace the therapies that the expert does, but to provide informative support.
- E2: "*Did you consider the standard of the electronic medical record for the design of this program? There are facial expressions that do not universally correspond to an emotion, the same gesture (raising the eyebrow) can be a sign of several emotions depending on the person, how can the bias be reduced? I see it as useful for research and educational contexts rather than psychotherapeutic. Excellent initiative.*"
    - o Our answer: Facial expressions can be mistaken for emotions, but micro-expressions are more difficult to mistake for emotions. The model recognizes the most descriptive patterns of the face, such as micro-expressions. It is possible to improve recognition for people without facial expressions with physiological sensors.

## 4. Limitations

Qualitative evaluation with experts considering its experience and treated patients is valuable for others psychologists and psychiatrists, also for mental health institutions. But the sample size of experts is no enough to test a hypothesis of interest.

The web application achieves real-time emotion recognition, even under the use of limited resources in devices. But it has a communication limitation, the voice IP module is not included in the current web application. Therefore, now, psychologists and psychiatrists must make a phone call to their patients.

## 5. Conclusions

There is an imminent need for psychological and psychiatric intervention over society is emotionally affected by the circumstances of the COVID-19 pandemic. Our web application with Facial Emotion Recognition (FER) could support this need.
We presented the ResMoNet model for FER, which was mainly based on the depthwise separable convolutions presented in MobileNet and the residuality advantage proved in IDNN, EDNN, and PeleeNet, with a major adjustment in the Residual Block. The model architecture generates a smaller number of parameters and mult-add operations. Also, the trained model reduced the time to recognize emotion and the main memory utilization.

A web application for psychologists and psychiatrists is proposed, which provides access to medical records, plots of patient emotions (anger, disgust, fear, happiness, sadness, surprise, and neutral), and a register of medical patient activity. Five experts evaluated the usability (73.8 of 100) and utility (3.94 of 5) of this tool as good.

For evidence-based emotional disorders strategies, this proposal for psychological support service can be integrated into the new platform as suggested by [47]. Moreover, voice IP modules need to be considered in a platform with low data transfer.

**Declarations**
This work is supported by the FORDECYT 296737 project "Consorcio en Inteligencia Artificial" for the publication of this work.
We declare no conflict of interests or competing interests.
The psychiatrists and psychologists consented to participate in the experts' opinion questionnaires.

*Appendix 1: Questionnaires*

Usability questionnaire.
1. I think I can use this system frequently.
2. I think the system is unnecessarily complex.
3. I think the system is easy to use.
4. I think I could need help from technical staff to use the system.
5. I think many functions of the system are well integrated
6. I think there are many inconsistencies in the system
7. Other people could learn to use this system quickly
8. I think this system is uncomfortable to use
9. I feel so sure to use this system
10. I need to learn many things before I use this system

Utility questionnaire:

11. I found the system helps to understand better the emotions of patients on the COVID-19 pandemic.
12. I think the system helps to record enough information from patients.
13. I found the system is useful for institutes of mental health on time of COVID-19 pandemic.
14. I think the system can be useful for other psychiatrists or psychologists.

Opened questions:
15. The circumstances in times of the COVID-19 pandemic have affected the mental health of society. How many patients have you treated in the time of the COVID-19 pandemic?
16. Suggestions and comments to the web application.